\documentclass{article}

\PassOptionsToPackage{numbers, sort&compress}{natbib}

\usepackage[dblblindworkshop,final]{neurips_2025}

\usepackage{multirow}
\usepackage{microtype}
\usepackage{amsmath}
\usepackage{amssymb}
\usepackage{booktabs}
\usepackage{graphicx}
\usepackage{enumitem}
\usepackage{xcolor}
\usepackage{tabularx}
\usepackage{subcaption}

\usepackage[detect-all,separate-uncertainty = true]{siunitx}
\makeatletter
\renewcommand{\paragraph}{%
  \@startsection{paragraph}{4}%
  {\z@}{1.5ex \@plus 1ex \@minus .2ex}{-0.5em}%
  {\normalfont\normalsize\bfseries}%
}
\makeatother

\setlist{nosep}

\definecolor{cvprblue}{rgb}{0.21,0.49,0.74}
\usepackage[pagebackref,breaklinks,colorlinks,allcolors=cvprblue]{hyperref}
\usepackage[capitalise]{cleveref}

\title{Mind the (Data) Gap: Evaluating Vision Systems in Small Data Applications}

\author{Samuel Stevens\\
The Ohio State University\\
{\tt\small stevens.994@osu.edu}
\and
S M Rayeed\\
Rensselaer Polytechnic Institute\\
{\tt\small rayees@rpi.edu}
\and
Jenna Kline\\
The Ohio State University\\
{\tt\small kline.377@osu.edu}
}

\begin{document}
\maketitle
\begin{abstract}
The practical application of AI tools for specific computer vision tasks relies on the ``small-data regime'' of hundreds to thousands of labeled samples. 
This small-data regime is vital for applications requiring expensive expert annotations, such as ecological monitoring, medical diagnostics or industrial quality control.
We find, however, that computer vision research has ignored the small data regime as evaluations increasingly focus on zero- and few-shot learning.
We use the Natural World Tasks (NeWT) benchmark to compare multi-modal large language models (MLLMs) and vision-only methods across varying training set sizes. 
MLLMs exhibit early performance plateaus, while vision-only methods improve throughout the small-data regime, with performance gaps widening beyond 10 training examples. 
We provide the first comprehensive comparison between these approaches in small-data contexts and advocate for explicit small-data evaluations in AI research to better bridge theoretical advances with practical deployments.
\end{abstract}

\section{Introduction}

\begin{figure}[t]
    \centering
    \begin{subfigure}[t]{0.49\textwidth}
        \centering
        \includegraphics[width=\linewidth]{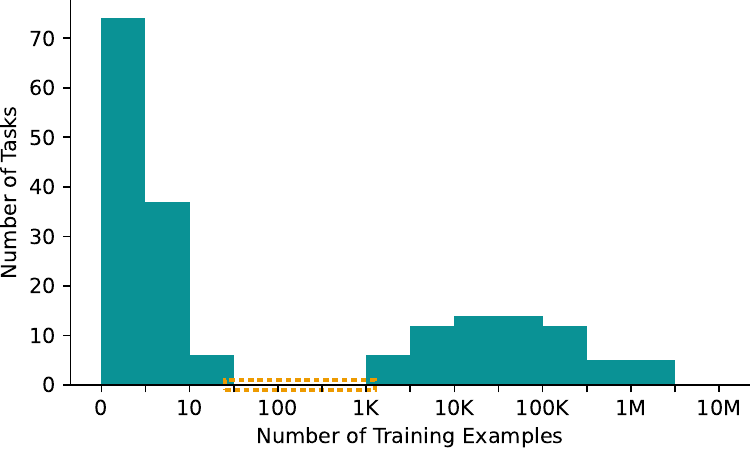}
        \phantomcaption{}\label{fig:tasks}
    \end{subfigure}%
    \hfill
    \centering
    \begin{subfigure}[t]{0.49\textwidth}
        \centering
        \includegraphics[width=\linewidth]{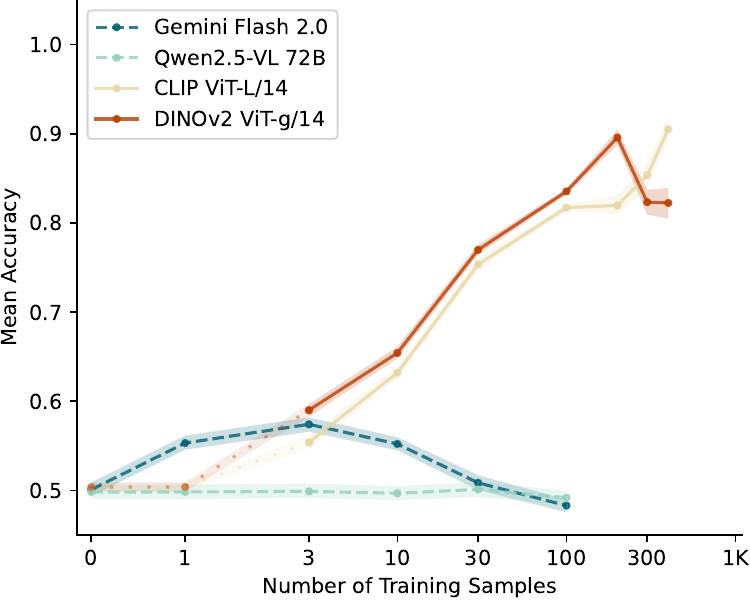}
        \phantomcaption{}\label{fig:hook}
    \end{subfigure}
    \caption{\textbf{Left:} Unique evaluation tasks used in recent language and vision research \citep{google2024gemini1.5,radford2021clip,oquab2023dinov2,zhai2023siglip,anthropic2024sonnet35,microsoft2024phi4,tschannen2025siglip2,meta2024llama32,olmo2024olmo2,lambert2024tulu3,fini2024aimv2,bardes2024vjepa,google2025gemma3} summarized by the number of training samples per task.
    Note how few evaluations use between \num{10} and \num{1000} labeled training samples.
    We collect this data manually.
    \textbf{Right:} Mean NeWT task performance as a function of number of labeled examples for multimodal large language models (MLLMs) and vision-only models combined with support vector machines (SVMs).
    MLLMs leverage labeled examples by including additional labeled examples in the prompt (few-shot prompting).
    Vision models leverage training examples by fitting an SVM to frozen image embeddings.
    Vision models with SVMs improve with additional training data and consistently outperform MLLMs with \num{10} or more labeled samples.
    Note the log scale for training data.
    Shaded areas indicate bootstrapped \num{95}\% confidence intervals.}
\end{figure}

AI research has increasingly favored evaluating new methods primarily through zero-shot and few-shot benchmarks~\cite{brown2020gpt3,radford2021clip,bommasani2021opportunities,hendrycks2020mmlu,yue2023mmmu}.
This evaluation approach is driven by the compelling promise of strong generalization with minimal examples. 
However, this focus on zero—and few-shot learning neglects a pervasive and essential scenario: the \textit{small-data regime}, characterized by datasets containing roughly dozens to a few thousand labeled samples (see \cref{fig:tasks}). 
This regime is critical for numerous real-world applications where extensive labeled data collection remains costly and challenging,  such as ecological monitoring \citep{van2018inat2017,beery2018recognition}, medical diagnostics \citep{esteva2017dermatologist}, and industrial quality control \citep{wang2018deep}.
Our community's decreased attention to rigorous small-data evaluations is a significant oversight. 
By optimizing primarily for zero-shot and few-shot performance, we risk developing methods ill-suited for practical scenarios where moderate data availability is typical. 
To address this gap, evaluations of new methods should explicitly include small-data assessments.

To systematically evaluate the small-data regime, we use the Natural World Tasks \citep[NeWT;][]{van2021inat2021} benchmark, which is specifically designed for challenging fine-grained ecological classification tasks requiring expert annotation. 
Using this benchmark, we compare multimodal large language models (MLLMs, \`a la Qwen2.5-VL or Gemini Flash 2.0) and vision transformers (ViTs) across varying training set sizes.
Using ecological classification tasks as representative test cases, we analyze model performance and scaling behavior. 
Our findings highlight significant limitations of current MLLMs, notably early performance plateaus, in contrast to sustained performance improvements observed in vision-only methods as dataset sizes increase within the small-data regime (see \cref{fig:hook}).
While our study utilizes ecological tasks as a convenient testbed due to the availability of the NeWT benchmark in the small-data regime, our argued position extends beyond ecology to the broader field of computer vision applications where limited labeled data is a common constraint.
In this work, we:
\begin{enumerate}
\item{Emphasize the critical-but-neglected small-data regime and advocate for its inclusion in AI research benchmarks.}
\item{Conduct the first comparison of foundation models versus vision-only methods within the small-data regime.}
\end{enumerate}
This work profiles performance patterns across model types and data scales.
While our findings can inform new research directions, we avoid model selection advice as optimal approaches depend on application-specific constraints. 
We aim to present empirical evidence highlighting the need for small-data evaluations in AI research.

\section{Background \& Related Work}\label{sec:related-work}

We highlight a gap in evaluation practices,  discuss trends and visually summarize the \textit{small-data gap} in \cref{fig:tasks}.

\paragraph{Evaluation Trends in AI Research.} 
Recent computer vision methods \citep{radford2021clip,oquab2023dinov2}, and (multimodal) large language models \citep[(M)LLMs;][]{openai2024gpt4o,anthropic2024sonnet35,google2025gemini2,bai2025qwen25vl} primarily evaluate performance using zero-shot or few-shot benchmarks \citep{hendrycks2020mmlu,suzgun2022bigbenchhard,yue2023mmmu}. 
These benchmarks emphasize generalization with extremely limited examples, reflecting a trend toward model robustness with minimal fine-tuning.

\paragraph{Evaluation Trends in Ecological Computer Vision.}
While ecological computer vision has begun adopting multimodal and foundation models for tasks such as species identification \citep{van2018inat2017,van2021inat2021}, the evaluations still frequently rely on fixed data splits or zero/few-shot scenarios. For instance, ecological benchmarks such as iNat2021 \citep{van2021inat2021} or iWildCam \citep{koh2021wilds,beery2021iwildcam} evaluate systems on \num{10}K+ labeled examples without systematically exploring performance scaling within moderate-sized training sets.
Recent specialized foundation models \citep{stevens2024bioclip,gong2024clibd,sastry2024taxabind} demonstrate interest in domain-specific representations, yet small-data evaluations remain uncommon.

\paragraph{Small-Data Gap}
Despite the practical importance of evaluating methods with tens to thousands of labeled examples, a regime typical in real-world ecological, medical, and industrial applications \citep{brigato2021close,kraljevski2023machine}, current methods research neglect systematic evaluation at these scales (see \cref{fig:tasks}). 
This oversight is significant: models optimized solely for zero- or few-shot benchmarks risk poor alignment with realistic deployments where moderate labeled datasets are both common and crucial.

Our work addresses this critical gap, systematically comparing foundation models (MLLMs) with vision-only methods, explicitly highlighting performance characteristics in the neglected small-data regime.

\begin{figure*}[t]
    \includegraphics[width=\textwidth]{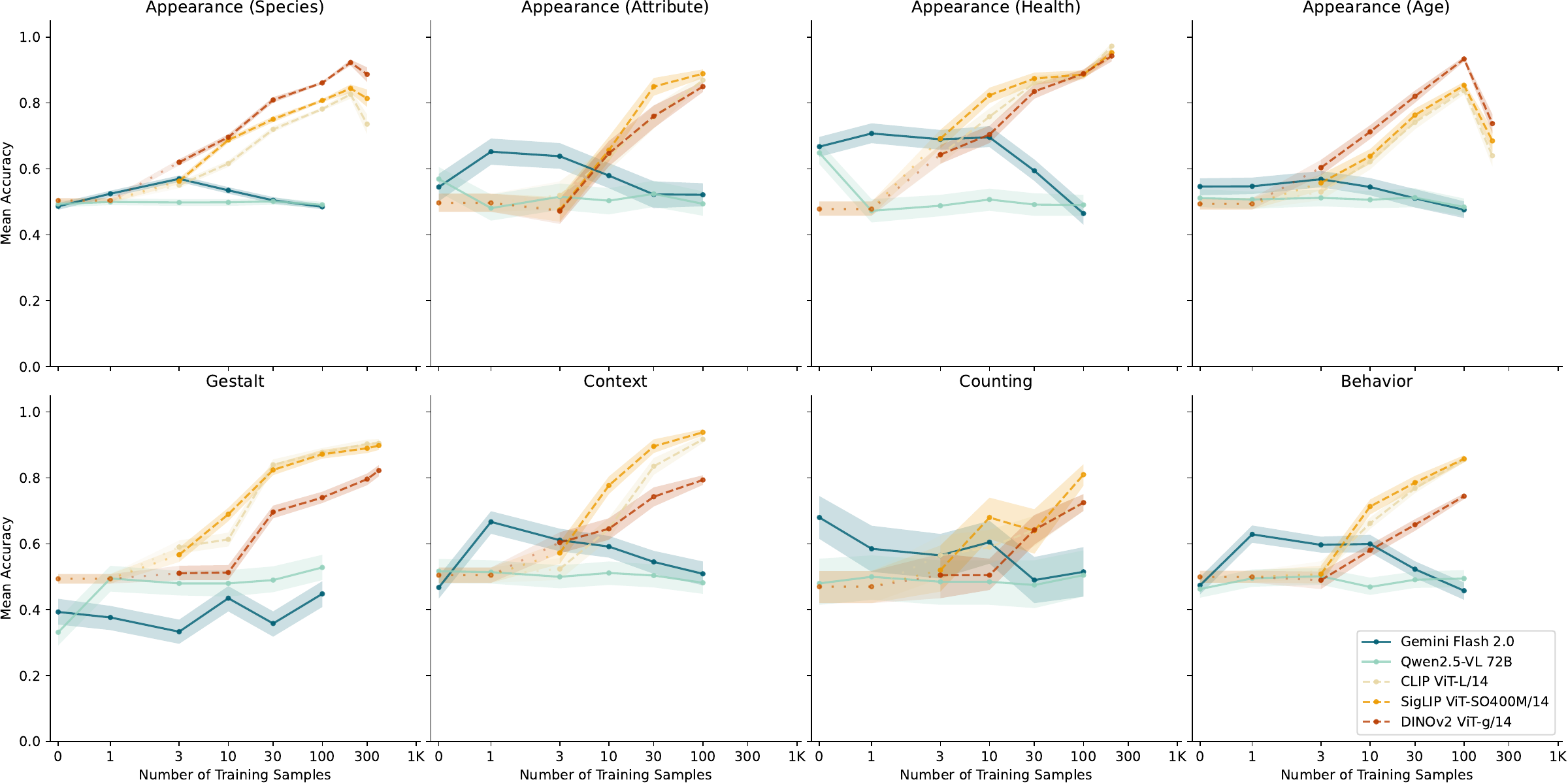}
    \caption{Performance scaling across NeWT's \citep{van2021inat2021} eight task clusters as a function of number of labeled examples. 
    Each panel corresponds to one task \textit{cluster} (species, attributes, health, ages, gestalt, context, counting, behavior; clusters contain more than one task). 
    Lines depict representative multimodal large language models (MLLMs: Gemini Flash 2.0, Qwen2.5-VL 72B) and vision encoders (CLIP ViT-L/14, DINOv2 ViT-g/14, SigLIP ViT-SO400M/14).
    Shaded regions represent \num{95}\% bootstrapped confidence intervals. 
    MLLMs exhibit early performance plateaus compared to sustained improvements seen in vision encoders combined with SVMs as the number of labeled examples.
    We cannot fit SVMs without at least one labeled example per class; we simulate random chance for \num{0} and \num{1} labeled examples.}\label{fig:main}
\end{figure*}

\section{Methodology}\label{sec:methodology}

To highlight the overlooked small-data regime in AI research, we introduce a rigorous experimental framework comparing multi-modal large language models (MLLMs) and vision transformers combined with traditional machine learning approaches specifically within this scenario. 
Unlike widely-studied zero-shot and few-shot benchmarks, our experiments explicitly target moderate data scales, ranging from tens to thousands of samples, to uncover non-obvious scaling behaviors and limitations of state-of-the-art methods. 
By providing detailed methodological guidance, we encourage researchers to adopt similar small-data evaluations, facilitating meaningful insights and practical recommendations for future method development.

We evaluate MLLMs and vision-only models with support vector machines (SVMs) as representative paradigms, as both have demonstrated strong performance across diverse visual tasks yet remain insufficiently characterized within the small-data regime. 
MLLMs have primarily been evaluated on zero-shot or few-shot benchmarks, leaving their performance unclear when moderate quantities of labeled data are available. 
Conversely, traditional vision encoder-based methods, which explicitly leverage fine-tuning or transfer learning, might exhibit fundamentally different scaling behaviors. 
Our methodology thus aims to elucidate previously unobserved contrasts and limitations by directly comparing these approaches within this under-explored setting.

Specifically, we evaluate multiple state-of-the-art MLLMs (e.g., Gemini Flash, Qwen2.5-VL) alongside vision encoders (e.g., DINOv2, CLIP variants) paired with SVM-based classifiers on diverse ecological datasets from the NeWT benchmark. 
We systematically vary the number of labeled examples and apply standard prompting and parsing procedures to rigorously characterize model behaviors and scaling trends.
\cref{sec:methodology} contain our prompts and pseudo-code for parsing responses.

\paragraph{Tasks}
We evaluate models on the NeWT benchmark \citep{van2021inat2021}.
NeWT contains \num{164} ecologically-motivated binary classification tasks, each with \num{200} to \num{400} labeled examples.
Tasks are grouped into eight clusters: species, attributes, health, ages, gestalt, contexts, counting and behavior.
See both \cref{app:newt} and the original text for additional details.

\paragraph{Multimodal Large Language Models (MLLMs):} We evaluate Gemini Flash 2.0, Gemini Flash 1.5 8B, Qwen2-VL 7B and Qwen2.5-VL 72B.
At the time of writing, these models represent the best tradeoff between cost and performance on a variety of benchmarks.

\paragraph{Vision Encoders with SVMs:} We extract features from DINOv2, CLIP, and SigLIP, three popular pre-trained, general-domain vision encoders. 
We test ViT-B, ViT-L, and ViT-H variants.
Per NeWT's original methodology, we exclusively use SVMs as our binary classifier on top of dense vision model features; SVM hyperparameters are tuned using \texttt{scikit-learn}’s cross-validation grid search \citep{scikit-learn}.

\paragraph{Labeled Examples}
To analyze performance scaling, we train models on different amounts of labeled examples. 
We define near-logarithmically scaled training subsets with sample sizes of \num{0}, \num{1}, \num{3}, \num{10}, \num{30}, \num{100}, \num{300}, and \text{all} examples.
Labeled examples are sampled uniformly and we ensure that there is at least one example per class when there are two or more examples.

\paragraph{Evaluation}
For all tasks, we compute bootstrapped confidence intervals by resampling test sets \num{1000} times with replacement and reporting the \num{95}\% confidence interval.
MLLM responses are parsed using deterministic regex-based extraction. 
If multiple species are listed, we take the first species mentioned.

\section{Results}\label{sec:results}

Our experiments reveal distinct performance characteristics between MLLMs and vision-only methods across the small-data spectrum.
Specific to ecological computer vision tasks, several notable patterns emerge that challenge conventional assumptions.

\subsection{Scaling Data}\label{sec:data-efficiency}

As shown in \cref{fig:main}, MLLMs and vision-only methods exhibit fundamentally different scaling behaviors as the number of labeled examples increases. 
MLLMs demonstrate rapid initial gains with very few examples (1-3) but consistently reach performance plateaus after 10-30 examples across most task clusters. 
In contrast, leveraging SVMs with vision transformers \citep[ViTs;][]{dosovitskiy2020image} show continuous, near-logarithmic improvement throughout the entire small-data regime, with no evidence of plateauing.
This scaling disparity results in a widening performance gap as dataset size increases.

\subsection{Scaling Models}\label{sec:model-efficiency}

\begin{figure}[t]
    \centering
    \begin{subfigure}[t]{0.49\textwidth}
        \centering
        \includegraphics[width=\linewidth]{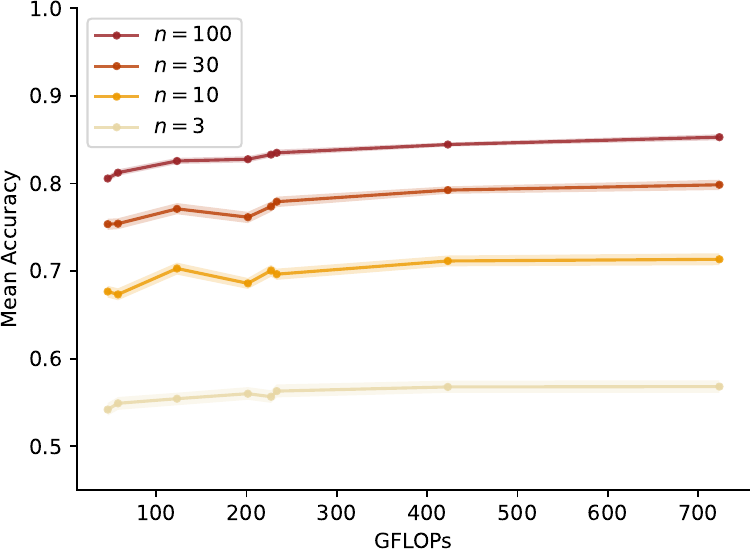}
        \phantomcaption{}\label{fig:flops}
    \end{subfigure}%
    \hfill
    \centering
    \begin{subfigure}[t]{0.49\textwidth}
        \centering
        \includegraphics[width=\linewidth]{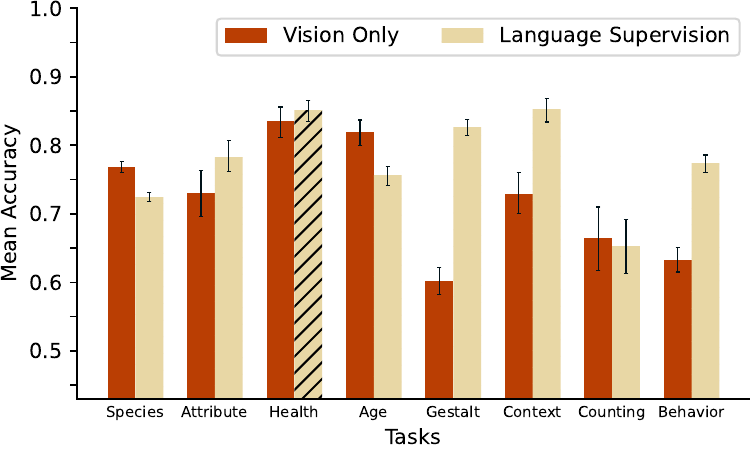}
        \phantomcaption{}\label{fig:pre-training}
    \end{subfigure}
    \vspace{-12pt}
    \caption{\textbf{Left:} 
    Vision model performance with respect to inference FLOPs and number of labeled examples ($n$).
    SigLIP \cite{zhai2023siglip} released eight different pre-trained transformers with varying model sizes (ViT-B/16, ViT-L/16 and ViT-SO400M/14) and image sizes ($224\times224$, $256\times256$, $384\times384$, and $512\times512$); we unify these axes with FLOPs/image.
    We find that increasing the number of labeled examples is more effective than increasing the model size; a $10\times$ increase in labeled examples outperforms a $10\times$ increase in FLOPs.
    \textbf{Right:} 
    Comparing vision model pre-training on performance across the eight task clusters in NeWT for \num{30} labeled examples with ViT-L models.
    Black error bars indicate bootstrapped \num{95}\% confidence intervals.
    Vision-only pre-training \citep[DINOv2;][]{oquab2023dinov2} outperforms language-supervised pre-training \citep[CLIP and SigLIP;][]{radford2021clip,zhai2023siglip} on `Species' and 'Age' tasks, both of which are fine-grained classification tasks.
    We observe that language supervision leads to large improvements on `Gestalt' and `Behavior' tasks, both of which require semantic reasoning.
    These conclusions hold for other numbers of labeled examples; see \cref{app:pretraining} for additional results.}
\end{figure}

Adding parameters to large language models demonstrates consistent improvement \citep{kaplan2020scaling,hoffmann2022chinchilla,wei2022emergent}.
Our analysis reveals a different pattern for vision models in ecological tasks.
As \cref{fig:flops} illustrates, increasing computational resources yields diminishing returns compared to simply adding more labeled examples. 
Even as we scale SigLIP models across model and image sizes from \num{45} to \num{700}+ GFLOPs, accuracy improvements remain modest, with a $10\times$ increase in labeled samples consistently outperforming a $10\times$ increase in computational capacity. 
This challenges the dominant ``bigger is better'' paradigm in recent AI research \citep{mckenzie2023inverse,wei2022inverse}.

Several factors could explain this difference from language model scaling properties.
The emergence threshold for vision models might occur at parameter counts beyond our experimental range \citep{sun2023eva,fang2024eva2,sun2024eva18b,dehghani2023scaling}.
Pretraining methodology differences are significant—vision models employ diverse objectives (contrastive, self-supervised, supervised) compared to the converged next-token prediction approach in language.
Our findings indicate that for ecological computer vision tasks within the small-data regime, prioritizing data collection provides more reliable performance improvements than scaling computational resources alone.

\subsection{Pre-Training Supervision}\label{sec:pretraining}

We use these results to reveal distinct performance patterns between vision-only and language-supervised pre-training approaches across ecological task clusters (\cref{fig:pre-training}). These differences underscore how pre-training objectives fundamentally influence a model's capabilities.

Vision-only pre-training (DINOv2) significantly outperforms language-supervised approaches (CLIP, SigLIP) on fine-grained visual discrimination tasks, specifically `Species' and `Age' classification. This advantage likely stems from DINOv2's self-supervised training objective, which builds rich hierarchical representations through local-to-global correspondence without language constraints. Such representations excel at capturing subtle morphological differences crucial for taxonomic identification and age determination tasks.

Conversely, language-supervised pre-training demonstrates substantial advantages in tasks requiring semantic understanding and contextual reasoning, notably in `Gestalt', `Context', and `Behavior' clusters, where models must recognize abstract visual concepts like image quality or animal activities. 
This suggests that image-text learning provides semantic grounding that pure vision models lack. 

These observed differences between pre-training methods are consistent across training set sizes (\cref{app:pretraining}), suggesting fundamental differences in learned representations, which is theoretically supported by recent work in interpreting vision models \citep{stevens2025saev,thasarathan2025universal}.

\section{Conclusion \& Future Work}\label{sec:conclusion}

Our systematic evaluation of the small-data regime reveals distinct performance patterns: vision-only systems exhibit sustained improvement while MLLMs demonstrate early performance plateaus beyond 10-30 labeled examples, suggesting that prompting struggles to learn nuanced patterns beyond a critical threshold of examples \citep{jiang2024many,zhang2024out}.
Our findings underscore the critical importance of evaluating AI methods explicitly within the small-data regime, an evaluation practice largely overlooked in current work despite its relevance to real applications.
By highlighting this evaluation gap, we hope to encourage more comprehensive benchmarking practices that better reflect the diverse data contexts encountered in practice.


\clearpage

{
    \small
    \bibliographystyle{ieeenat_fullname}
    \bibliography{main}
}


\clearpage

\appendix

\section*{Appendices}

\begin{enumerate}
    \item{\cref{app:methodology-details}: Methodology details for \cref{sec:methodology}.}
    \item{\cref{app:newt}: NeWT details for \cref{sec:methodology}.}
    \item{\cref{app:pretraining}: Additional results for \cref{sec:pretraining}.}
    \item{\cref{app:all-results}: All raw results.}
\end{enumerate}

\section{Methodology Details}\label{app:methodology-details}

We evaluate two families of models: (1) multimodal large language models (MLLMs) and (2) vision encoders with machine learning classifiers.

\paragraph{MLLMs:} We test the following models via API access: Gemini Flash 2.0, Gemini Flash 1.5 8B, Qwen2-VL 7B and Qwen2.5-VL 72B. 

\paragraph{Vision Encoders:} We extract image embeddings from DINOv2, CLIP, and SigLIP and include ViT-B, ViT-L, and ViT-H variants. 

\paragraph{Image Preprocessing:}  
All images are resized so that the smaller side is 224 pixels, then center cropped to $224\times224$. 
We normalize images using the ImageNet mean and standard deviation for all models. 
No additional augmentations (e.g., cropping, flipping) are applied for inference.
Images are not modified before being sent to MLLMs.

\paragraph{Sampling:}
Subsets of labeled examples are sampled uniformly from the full training split with the following sample sizes: \num{0}, \num{1}, \num{3}, \num{10}, \num{30}, \num{100}, and \num{300} where applicable. 

\paragraph{Prompting and Parsing:}
All the tasks in NeWT are binary classification tasks. We use the following template: ``\texttt{What is this a picture of, '\{a\}' or '\{b\}'? Respond with your answer in bold.}'' where \texttt{\{a\}} and \texttt{\{b\}} are replaced with the two classes, with a random order.
MLLM responses are parsed using regex-based extraction, using character-based distance to pick the classname closest to whatever bold text is first found in the response.

\paragraph{Classifier Hyperparameters}
We perform a search over the following hyperparameter distribution:
\begin{itemize}
    \item{C: log-uniform distribution from $10^{-3}$ to $10^1$.}
    \item{Kernel: one of RBF, linear, sigmoid or cubic}
    \item{Kernel coefficient: log-uniform distribution from $10^{-4}$ to $10^{-3}$. Ignored for linear kernel.}
\end{itemize}
We sample 100 models and evaluate with $5$-fold cross-validation over the training set.

\paragraph{Bootstrapped Confidence Intervals}
For all evaluation metrics, we report bootstrapped confidence intervals:
\begin{itemize}
    \item We resample the test set \num{1000} times with replacement.
    \item The mean accuracy is computed for each resampled test set.
    \item The 95\% confidence interval is reported.
\end{itemize}

\paragraph{Compute Infrastructure:}  
ViT-based inference is batched on NVIDIA A6000 GPUs to maximize GPU memory efficiency. 
API-based MLLM inference is conducted on cloud platforms.


\section{Background on NeWT}\label{app:newt}

Natural World Tasks \citep[NeWT;][]{van2021inat2021} is a collection of \num{164} binary classification tasks that go beyond species classification.
The tasks are manually curated with a uniform distribution of positive and negative examples (so accuracy is an appropriate metric).

\section{Pre-Training Supervision}\label{app:pretraining}

\begin{figure*}[t]
    \includegraphics[width=\textwidth]{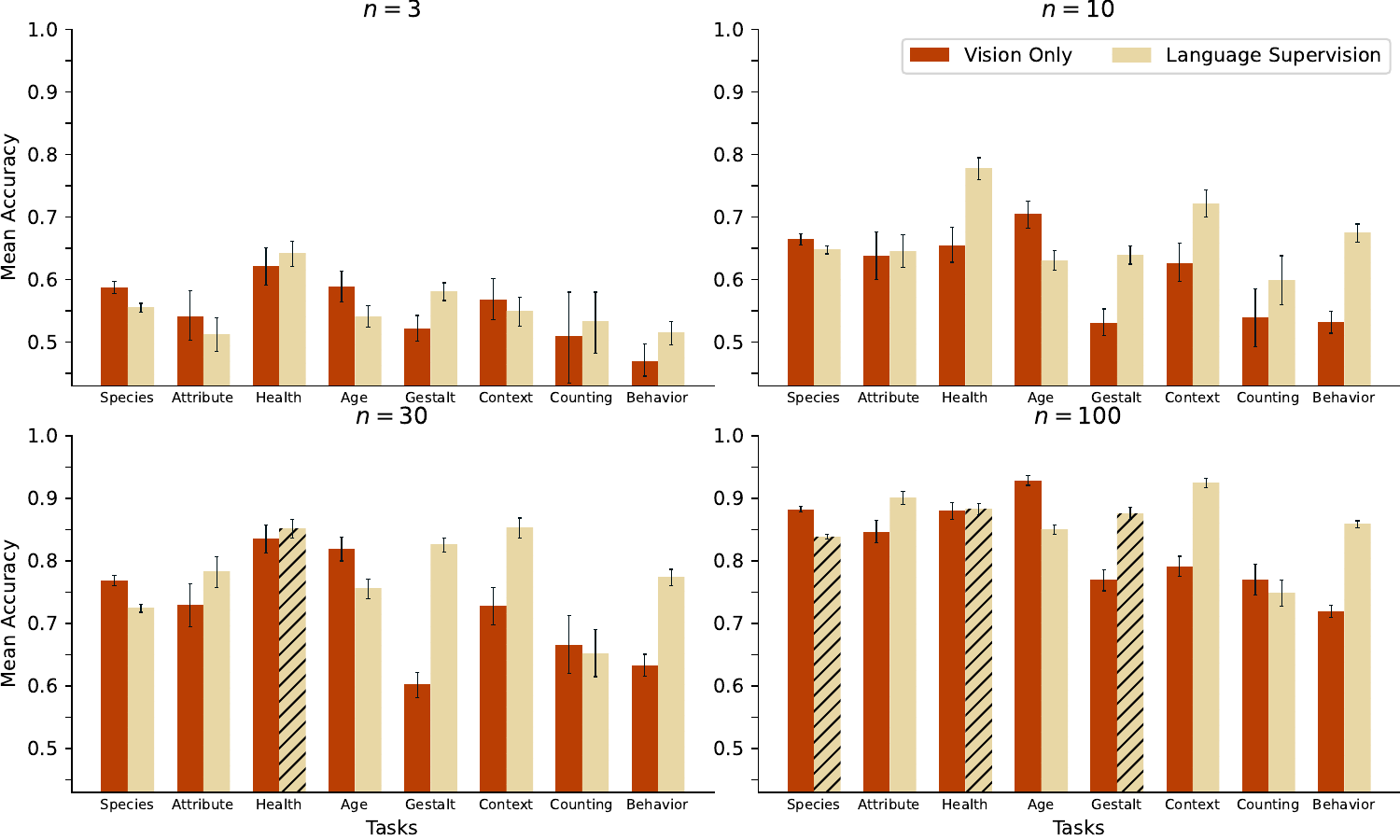}
    \caption{Comparing vision model pre-training on performance across the eight task clusters in NeWT for \num{3}, \num{10}, \num{30}, and \num{100} training samples with ViT-L models.}\label{fig:pretraining-all}
\end{figure*}

\cref{fig:pretraining-all} contains results for \num{3}, \num{10}, \num{30}, and \num{100} training samples, summarized by the ViT's pre-training objective.
Vision-only pre-training outperforms vision-language pre-training on `Species' and `Age' tasks, while vision-language pre-training outperforms vision-only pre-training on more semantic tasks (`Gestalt', `Context', and `Behavior').

\section{All Results}\label{app:all-results}

We include all results in \cref{tab:all-results-n0,tab:all-results-n1,tab:all-results-n3-species-attribute,tab:all-results-n3-health-age,tab:all-results-n3-gestalt-context,tab:all-results-n3-counting-behavior,tab:all-results-n10-species-attribute,tab:all-results-n10-health-age,tab:all-results-n10-gestalt-context,tab:all-results-n10-counting-behavior,tab:all-results-n30-species-attribute,tab:all-results-n30-health-age,tab:all-results-n30-gestalt-context,tab:all-results-n30-counting-behavior,tab:all-results-n100-species-attribute,tab:all-results-n100-health-age,tab:all-results-n100-gestalt-context,tab:all-results-n100-counting-behavior}.
These results will also be made available in a machine-readable format.

\begin{table*}[t]
\small
\centering
\begin{tabular}{lllrrr}
\toprule
Task Cluster & Task Subcluster & Model & Train & Mean & Confidence Interval \\ \midrule

Appearance & Species & Gemini Flash 2.0 & $0$ & $0.49$ & $[0.48, 0.50]$ \\

Appearance & Species & Gemini Flash 1.5 8B & $0$ & $0.50$ & $[0.49, 0.51]$ \\

Appearance & Species & Qwen2.5-VL 72B & $0$ & $0.49$ & $[0.48, 0.50]$ \\

Appearance & Species & Qwen2-VL 7B & $0$ & $0.50$ & $[0.49, 0.51]$ \\\midrule

Appearance & Attribute & Gemini Flash 2.0 & $0$ & $0.54$ & $[0.51, 0.59]$ \\

Appearance & Attribute & Gemini Flash 1.5 8B & $0$ & $0.60$ & $[0.56, 0.64]$ \\

Appearance & Attribute & Qwen2.5-VL 72B & $0$ & $0.57$ & $[0.53, 0.61]$ \\
\bottomrule
\end{tabular}
\caption{
All results for $0$ training samples.
}\label{tab:all-results-n0}
\end{table*}
\begin{table*}[t]
\small
\centering
\begin{tabular}{lllrrr}
\toprule
Task Cluster & Task Subcluster & Model & Train & Mean & Confidence Interval \\ \midrule

Appearance & Species & Gemini Flash 2.0 & $1$ & $0.52$ & $[0.52, 0.53]$ \\

Appearance & Species & Gemini Flash 1.5 8B & $1$ & $0.54$ & $[0.53, 0.55]$ \\

Appearance & Species & Qwen2.5-VL 72B & $1$ & $0.50$ & $[0.49, 0.51]$ \\

Appearance & Species & Qwen2-VL 7B & $1$ & $0.50$ & $[0.49, 0.51]$ \\\midrule

Appearance & Attribute & Gemini Flash 2.0 & $1$ & $0.65$ & $[0.61, 0.69]$ \\

Appearance & Attribute & Gemini Flash 1.5 8B & $1$ & $0.68$ & $[0.64, 0.71]$ \\

Appearance & Attribute & Qwen2.5-VL 72B & $1$ & $0.48$ & $[0.44, 0.52]$ \\
\bottomrule
\end{tabular}
\caption{
All results for $1$ training sample.
}\label{tab:all-results-n1}
\end{table*}
\begin{table*}[t]
\small
\centering
\begin{tabular}{lllrrr}
\toprule
Task Cluster & Task Subcluster & Model & Train & Mean & Confidence Interval \\ \midrule

Appearance & Species & Gemini Flash 2.0 & $3$ & $0.57$ & $[0.56, 0.58]$ \\

Appearance & Species & Gemini Flash 1.5 8B & $3$ & $0.53$ & $[0.52, 0.54]$ \\

Appearance & Species & SigLIP ViT-B/16 (512px) & $3$ & $0.56$ & $[0.55, 0.57]$ \\

Appearance & Species & SigLIP ViT-L/16 (256px) & $3$ & $0.56$ & $[0.55, 0.57]$ \\

Appearance & Species & SigLIP ViT-L/16 (384px) & $3$ & $0.57$ & $[0.56, 0.58]$ \\

Appearance & Species & SigLIP ViT-SO400M/14 & $3$ & $0.56$ & $[0.55, 0.57]$ \\

Appearance & Species & SigLIP ViT-SO400M/14 (384px) & $3$ & $0.56$ & $[0.56, 0.57]$ \\
\bottomrule
\end{tabular}
\caption{
All results for $3$ training samples for `Species' and `Attribute' tasks.
}\label{tab:all-results-n3-species-attribute}
\end{table*}
\begin{table*}[t]
\small
\centering
\begin{tabular}{lllrrr}
\toprule
Task Cluster & Task Subcluster & Model & Train & Mean & Confidence Interval \\ \midrule

Appearance & Health & Gemini Flash 2.0 & $3$ & $0.69$ & $[0.66, 0.72]$ \\

Appearance & Health & Gemini Flash 1.5 8B & $3$ & $0.71$ & $[0.68, 0.74]$ \\

Appearance & Health & SigLIP ViT-B/16 (512px) & $3$ & $0.59$ & $[0.56, 0.62]$ \\

Appearance & Health & SigLIP ViT-L/16 (256px) & $3$ & $0.64$ & $[0.61, 0.67]$ \\

Appearance & Health & SigLIP ViT-L/16 (384px) & $3$ & $0.69$ & $[0.66, 0.72]$ \\

Appearance & Health & SigLIP ViT-SO400M/14 & $3$ & $0.69$ & $[0.67, 0.72]$ \\

Appearance & Health & SigLIP ViT-SO400M/14 (384px) & $3$ & $0.73$ & $[0.70, 0.76]$ \\
\bottomrule
\end{tabular}
\caption{
All results for $3$ training samples for `Health' and `Age' tasks.
}\label{tab:all-results-n3-health-age}
\end{table*}
\begin{table*}[t]
\small
\centering
\begin{tabular}{lllrrr}
\toprule
Task Cluster & Task Subcluster & Model & Train & Mean & Confidence Interval \\ \midrule

Gestalt & - & Gemini Flash 2.0 & $3$ & $0.33$ & $[0.29, 0.37]$ \\

Gestalt & - & Gemini Flash 1.5 8B & $3$ & $0.45$ & $[0.41, 0.49]$ \\

Gestalt & - & SigLIP ViT-B/16 (512px) & $3$ & $0.61$ & $[0.59, 0.63]$ \\

Gestalt & - & SigLIP ViT-L/16 (256px) & $3$ & $0.57$ & $[0.55, 0.59]$ \\

Gestalt & - & SigLIP ViT-L/16 (384px) & $3$ & $0.56$ & $[0.54, 0.58]$ \\

Gestalt & - & SigLIP ViT-SO400M/14 & $3$ & $0.57$ & $[0.55, 0.59]$ \\

Gestalt & - & SigLIP ViT-SO400M/14 (384px) & $3$ & $0.57$ & $[0.54, 0.59]$ \\
\bottomrule
\end{tabular}
\caption{
All results for $3$ training samples for `Gestalt' and `Context' tasks.
}\label{tab:all-results-n3-gestalt-context}
\end{table*}
\begin{table*}[t]
\small
\centering
\begin{tabular}{lllrrr}
\toprule
Task Cluster & Task Subcluster & Model & Train & Mean & Confidence Interval \\ \midrule

Counting & - & Gemini Flash 2.0 & $3$ & $0.56$ & $[0.50, 0.63]$ \\

Counting & - & Gemini Flash 1.5 8B & $3$ & $0.61$ & $[0.54, 0.68]$ \\

Counting & - & SigLIP ViT-B/16 (512px) & $3$ & $0.49$ & $[0.42, 0.56]$ \\

Counting & - & SigLIP ViT-L/16 (256px) & $3$ & $0.50$ & $[0.43, 0.57]$ \\

Counting & - & SigLIP ViT-L/16 (384px) & $3$ & $0.50$ & $[0.43, 0.56]$ \\

Counting & - & SigLIP ViT-SO400M/14 & $3$ & $0.52$ & $[0.45, 0.59]$ \\

Counting & - & SigLIP ViT-SO400M/14 (384px) & $3$ & $0.52$ & $[0.45, 0.59]$ \\
\bottomrule
\end{tabular}
\caption{
All results for $3$ training samples for `Counting' and `Behavior' tasks.
}\label{tab:all-results-n3-counting-behavior}
\end{table*}
\begin{table*}[t]
\small
\centering
\begin{tabular}{lllrrr}
\toprule
Task Cluster & Task Subcluster & Model & Train & Mean & Confidence Interval \\ \midrule

Appearance & Species & Gemini Flash 2.0 & $10$ & $0.53$ & $[0.53, 0.54]$ \\

Appearance & Species & Gemini Flash 1.5 8B & $10$ & $0.50$ & $[0.49, 0.52]$ \\

Appearance & Species & SigLIP ViT-B/16 (512px) & $10$ & $0.69$ & $[0.68, 0.70]$ \\

Appearance & Species & SigLIP ViT-L/16 (256px) & $10$ & $0.68$ & $[0.67, 0.69]$ \\

Appearance & Species & SigLIP ViT-L/16 (384px) & $10$ & $0.70$ & $[0.69, 0.70]$ \\

Appearance & Species & SigLIP ViT-SO400M/14 & $10$ & $0.69$ & $[0.68, 0.70]$ \\

Appearance & Species & SigLIP ViT-SO400M/14 (384px) & $10$ & $0.70$ & $[0.69, 0.71]$ \\
\bottomrule
\end{tabular}
\caption{
All results for $10$ training samples for `Species' and `Attribute' tasks.
}\label{tab:all-results-n10-species-attribute}
\end{table*}
\begin{table*}[t]
\small
\centering
\begin{tabular}{lllrrr}
\toprule
Task Cluster & Task Subcluster & Model & Train & Mean & Confidence Interval \\ \midrule

Appearance & Health & Gemini Flash 2.0 & $10$ & $0.70$ & $[0.66, 0.73]$ \\

Appearance & Health & Gemini Flash 1.5 8B & $10$ & $0.55$ & $[0.52, 0.59]$ \\

Appearance & Health & SigLIP ViT-B/16 (512px) & $10$ & $0.81$ & $[0.79, 0.84]$ \\

Appearance & Health & SigLIP ViT-L/16 (256px) & $10$ & $0.80$ & $[0.77, 0.82]$ \\

Appearance & Health & SigLIP ViT-L/16 (384px) & $10$ & $0.83$ & $[0.81, 0.85]$ \\

Appearance & Health & SigLIP ViT-SO400M/14 & $10$ & $0.82$ & $[0.80, 0.85]$ \\

Appearance & Health & SigLIP ViT-SO400M/14 (384px) & $10$ & $0.87$ & $[0.85, 0.89]$ \\
\bottomrule
\end{tabular}
\caption{
All results for $10$ training samples for `Health' and `Age' tasks.
}\label{tab:all-results-n10-health-age}
\end{table*}
\begin{table*}[t]
\small
\centering
\begin{tabular}{lllrrr}
\toprule
Task Cluster & Task Subcluster & Model & Train & Mean & Confidence Interval \\ \midrule

Gestalt & - & Gemini Flash 2.0 & $10$ & $0.43$ & $[0.40, 0.48]$ \\

Gestalt & - & Gemini Flash 1.5 8B & $10$ & $0.49$ & $[0.45, 0.53]$ \\

Gestalt & - & SigLIP ViT-B/16 (512px) & $10$ & $0.71$ & $[0.69, 0.73]$ \\

Gestalt & - & SigLIP ViT-L/16 (256px) & $10$ & $0.66$ & $[0.64, 0.69]$ \\

Gestalt & - & SigLIP ViT-L/16 (384px) & $10$ & $0.69$ & $[0.67, 0.71]$ \\

Gestalt & - & SigLIP ViT-SO400M/14 & $10$ & $0.69$ & $[0.67, 0.71]$ \\

Gestalt & - & SigLIP ViT-SO400M/14 (384px) & $10$ & $0.70$ & $[0.68, 0.72]$ \\
\bottomrule
\end{tabular}
\caption{
All results for $10$ training samples for `Gestalt' and `Context' tasks.
}\label{tab:all-results-n10-gestalt-context}
\end{table*}
\begin{table*}[t]
\small
\centering
\begin{tabular}{lllrrr}
\toprule
Task Cluster & Task Subcluster & Model & Train & Mean & Confidence Interval \\ \midrule

Counting & - & Gemini Flash 2.0 & $10$ & $0.60$ & $[0.53, 0.68]$ \\

Counting & - & Gemini Flash 1.5 8B & $10$ & $0.51$ & $[0.44, 0.57]$ \\

Counting & - & SigLIP ViT-B/16 (512px) & $10$ & $0.56$ & $[0.49, 0.62]$ \\

Counting & - & SigLIP ViT-L/16 (256px) & $10$ & $0.61$ & $[0.55, 0.68]$ \\

Counting & - & SigLIP ViT-L/16 (384px) & $10$ & $0.66$ & $[0.58, 0.71]$ \\

Counting & - & SigLIP ViT-SO400M/14 & $10$ & $0.68$ & $[0.61, 0.74]$ \\

Counting & - & SigLIP ViT-SO400M/14 (384px) & $10$ & $0.69$ & $[0.61, 0.74]$ \\
\bottomrule
\end{tabular}
\caption{
All results for $10$ training samples for `Counting' and `Behavior' tasks.
}\label{tab:all-results-n10-counting-behavior}
\end{table*}
\begin{table*}[t]
\small
\centering
\begin{tabular}{lllrrr}
\toprule
Task Cluster & Task Subcluster & Model & Train & Mean & Confidence Interval \\ \midrule

Appearance & Species & Gemini Flash 2.0 & $30$ & $0.50$ & $[0.49, 0.51]$ \\

Appearance & Species & Gemini Flash 1.5 8B & $30$ & $0.50$ & $[0.49, 0.51]$ \\

Appearance & Species & SigLIP ViT-B/16 (512px) & $30$ & $0.74$ & $[0.73, 0.75]$ \\

Appearance & Species & SigLIP ViT-L/16 (256px) & $30$ & $0.73$ & $[0.72, 0.74]$ \\

Appearance & Species & SigLIP ViT-L/16 (384px) & $30$ & $0.76$ & $[0.75, 0.76]$ \\

Appearance & Species & SigLIP ViT-SO400M/14 & $30$ & $0.75$ & $[0.74, 0.76]$ \\

Appearance & Species & SigLIP ViT-SO400M/14 (384px) & $30$ & $0.77$ & $[0.76, 0.78]$ \\
\bottomrule
\end{tabular}
\caption{
All results for $30$ training samples for `Species' and `Attribute' tasks.
}\label{tab:all-results-n30-species-attribute}
\end{table*}
\begin{table*}[t]
\small
\centering
\begin{tabular}{lllrrr}
\toprule
Task Cluster & Task Subcluster & Model & Train & Mean & Confidence Interval \\ \midrule

Appearance & Health & Gemini Flash 2.0 & $30$ & $0.59$ & $[0.56, 0.63]$ \\

Appearance & Health & Gemini Flash 1.5 8B & $30$ & $0.54$ & $[0.50, 0.57]$ \\

Appearance & Health & SigLIP ViT-B/16 (512px) & $30$ & $0.85$ & $[0.83, 0.87]$ \\

Appearance & Health & SigLIP ViT-L/16 (256px) & $30$ & $0.84$ & $[0.82, 0.86]$ \\

Appearance & Health & SigLIP ViT-L/16 (384px) & $30$ & $0.87$ & $[0.86, 0.89]$ \\

Appearance & Health & SigLIP ViT-SO400M/14 & $30$ & $0.87$ & $[0.85, 0.89]$ \\

Appearance & Health & SigLIP ViT-SO400M/14 (384px) & $30$ & $0.89$ & $[0.87, 0.91]$ \\
\bottomrule
\end{tabular}
\caption{
All results for $30$ training samples for `Health' and `Age' tasks.
}\label{tab:all-results-n30-health-age}
\end{table*}
\begin{table*}[t]
\small
\centering
\begin{tabular}{lllrrr}
\toprule
Task Cluster & Task Subcluster & Model & Train & Mean & Confidence Interval \\ \midrule

Gestalt & - & Gemini Flash 2.0 & $30$ & $0.36$ & $[0.32, 0.40]$ \\

Gestalt & - & Gemini Flash 1.5 8B & $30$ & $0.50$ & $[0.46, 0.54]$ \\

Gestalt & - & SigLIP ViT-B/16 (512px) & $30$ & $0.84$ & $[0.82, 0.85]$ \\

Gestalt & - & SigLIP ViT-L/16 (256px) & $30$ & $0.81$ & $[0.80, 0.83]$ \\

Gestalt & - & SigLIP ViT-L/16 (384px) & $30$ & $0.86$ & $[0.85, 0.87]$ \\

Gestalt & - & SigLIP ViT-SO400M/14 & $30$ & $0.82$ & $[0.81, 0.84]$ \\

Gestalt & - & SigLIP ViT-SO400M/14 (384px) & $30$ & $0.85$ & $[0.84, 0.87]$ \\
\bottomrule
\end{tabular}
\caption{
All results for $30$ training samples for `Gestalt' and `Context' tasks.
}\label{tab:all-results-n30-gestalt-context}
\end{table*}
\begin{table*}[t]
\small
\centering
\begin{tabular}{lllrrr}
\toprule
Task Cluster & Task Subcluster & Model & Train & Mean & Confidence Interval \\ \midrule

Counting & - & Gemini Flash 2.0 & $30$ & $0.49$ & $[0.43, 0.56]$ \\

Counting & - & Gemini Flash 1.5 8B & $30$ & $0.48$ & $[0.41, 0.55]$ \\

Counting & - & SigLIP ViT-B/16 (512px) & $30$ & $0.64$ & $[0.56, 0.70]$ \\

Counting & - & SigLIP ViT-L/16 (256px) & $30$ & $0.67$ & $[0.60, 0.73]$ \\

Counting & - & SigLIP ViT-L/16 (384px) & $30$ & $0.62$ & $[0.57, 0.67]$ \\

Counting & - & SigLIP ViT-SO400M/14 & $30$ & $0.64$ & $[0.57, 0.70]$ \\

Counting & - & SigLIP ViT-SO400M/14 (384px) & $30$ & $0.61$ & $[0.55, 0.69]$ \\
\bottomrule
\end{tabular}
\caption{
All results for $30$ training samples for `Counting' and `Behavior' tasks.
}\label{tab:all-results-n30-counting-behavior}
\end{table*}
\begin{table*}[t]
\small
\centering
\begin{tabular}{lllrrr}
\toprule
Task Cluster & Task Subcluster & Model & Train & Mean & Confidence Interval \\ \midrule

Appearance & Species & Gemini Flash 2.0 & $100$ & $0.48$ & $[0.48, 0.49]$ \\

Appearance & Species & Gemini Flash 1.5 8B & $100$ & $0.50$ & $[0.49, 0.51]$ \\

Appearance & Species & SigLIP ViT-B/16 (512px) & $100$ & $0.80$ & $[0.79, 0.80]$ \\

Appearance & Species & SigLIP ViT-L/16 (256px) & $100$ & $0.79$ & $[0.79, 0.80]$ \\

Appearance & Species & SigLIP ViT-L/16 (384px) & $100$ & $0.81$ & $[0.81, 0.82]$ \\

Appearance & Species & SigLIP ViT-SO400M/14 & $100$ & $0.81$ & $[0.80, 0.81]$ \\

Appearance & Species & SigLIP ViT-SO400M/14 (384px) & $100$ & $0.82$ & $[0.82, 0.83]$ \\
\bottomrule
\end{tabular}
\caption{
All results for $100$ training samples for `Species' and `Attribute' tasks.
}\label{tab:all-results-n100-species-attribute}
\end{table*}
\begin{table*}[t]
\small
\centering
\begin{tabular}{lllrrr}
\toprule
Task Cluster & Task Subcluster & Model & Train & Mean & Confidence Interval \\ \midrule

Appearance & Health & Gemini Flash 2.0 & $100$ & $0.46$ & $[0.43, 0.50]$ \\

Appearance & Health & Gemini Flash 1.5 8B & $100$ & $0.54$ & $[0.51, 0.58]$ \\

Appearance & Health & SigLIP ViT-B/16 (512px) & $100$ & $0.88$ & $[0.87, 0.89]$ \\

Appearance & Health & SigLIP ViT-L/16 (256px) & $100$ & $0.88$ & $[0.87, 0.89]$ \\

Appearance & Health & SigLIP ViT-L/16 (384px) & $100$ & $0.90$ & $[0.89, 0.90]$ \\

Appearance & Health & SigLIP ViT-SO400M/14 & $100$ & $0.89$ & $[0.87, 0.90]$ \\

Appearance & Health & SigLIP ViT-SO400M/14 (384px) & $100$ & $0.90$ & $[0.89, 0.91]$ \\
\bottomrule
\end{tabular}
\caption{
All results for $100$ training samples for `Health' and `Age' tasks.
}\label{tab:all-results-n100-health-age}
\end{table*}
\begin{table*}[t]
\small
\centering
\begin{tabular}{lllrrr}
\toprule
Task Cluster & Task Subcluster & Model & Train & Mean & Confidence Interval \\ \midrule

Gestalt & - & Gemini Flash 2.0 & $100$ & $0.45$ & $[0.41, 0.49]$ \\

Gestalt & - & Gemini Flash 1.5 8B & $100$ & $0.50$ & $[0.46, 0.54]$ \\

Gestalt & - & SigLIP ViT-B/16 (512px) & $100$ & $0.88$ & $[0.87, 0.89]$ \\

Gestalt & - & SigLIP ViT-L/16 (256px) & $100$ & $0.87$ & $[0.86, 0.89]$ \\

Gestalt & - & SigLIP ViT-L/16 (384px) & $100$ & $0.89$ & $[0.89, 0.90]$ \\

Gestalt & - & SigLIP ViT-SO400M/14 & $100$ & $0.87$ & $[0.86, 0.89]$ \\

Gestalt & - & SigLIP ViT-SO400M/14 (384px) & $100$ & $0.89$ & $[0.88, 0.90]$ \\
\bottomrule
\end{tabular}
\caption{
All results for $100$ training samples for `Gestalt' and `Context' tasks.
}\label{tab:all-results-n100-gestalt-context}
\end{table*}
\begin{table*}[t]
\small
\centering
\begin{tabular}{lllrrr}
\toprule
Task Cluster & Task Subcluster & Model & Train & Mean & Confidence Interval \\ \midrule

Counting & - & Gemini Flash 2.0 & $100$ & $0.52$ & $[0.45, 0.58]$ \\

Counting & - & Gemini Flash 1.5 8B & $100$ & $0.45$ & $[0.38, 0.52]$ \\

Counting & - & SigLIP ViT-B/16 (512px) & $100$ & $0.80$ & $[0.77, 0.83]$ \\

Counting & - & SigLIP ViT-L/16 (256px) & $100$ & $0.80$ & $[0.76, 0.83]$ \\

Counting & - & SigLIP ViT-L/16 (384px) & $100$ & $0.85$ & $[0.83, 0.88]$ \\

Counting & - & SigLIP ViT-SO400M/14 & $100$ & $0.81$ & $[0.78, 0.84]$ \\

Counting & - & SigLIP ViT-SO400M/14 (384px) & $100$ & $0.84$ & $[0.81, 0.87]$ \\
\bottomrule
\end{tabular}
\caption{
All results for $100$ training samples for `Counting' and `Behavior' tasks.
}\label{tab:all-results-n100-counting-behavior}
\end{table*}

\end{document}